\documentclass[letterpaper, 10 pt, conference]{ieeeconf}  

\IEEEoverridecommandlockouts                              

\usepackage{graphicx}
\usepackage{amsmath}
\usepackage{caption}
\usepackage{subfigure}
\usepackage[dvipsnames]{xcolor}

\title{\LARGE \bf
Predictive Temporal Attention on Event-based Video Stream for Energy-efficient Situation Awareness
}

\author{Yiming Bu, Jiayang Liu and Qinru Qiu\\Department of EECS, Syracuse University, Syracuse, USA\\\{ybu104, jliu206, qiqiu\}@syr.edu
\thanks{*This work is partially supported by the National Science
Foundation I/UCRC ASIC (Alternative Sustainable and Intelligent Computing) Center (CNS-1822165).}
}

\begin{document}

\maketitle
\thispagestyle{empty}
\pagestyle{empty}

\begin{abstract}

The Dynamic Vision Sensor (DVS) is an innovative technology that efficiently captures and encodes visual information in an event-driven manner. 
By combining it with event-driven neuromorphic processing, the sparsity in DVS camera output can result in high energy efficiency. However, similar to many embedded systems, the off-chip communication between the camera and processor presents a bottleneck in terms of power consumption. Inspired by the predictive coding model and expectation suppression phenomenon found in human brain, we propose a temporal attention mechanism to throttle the camera output and pay attention to it only when the visual events cannot be well predicted. The predictive attention not only reduces power consumption in the sensor-processor interface but also effectively decreases the computational workload by filtering out noisy events. 
We demonstrate that the predictive attention can reduce $46.7\%$ of data communication between the camera and the processor and reduce $43.8\%$ computation activities in the processor.

\end{abstract}

\section{INTRODUCTION}

While human eyes are sensitive to color information, certain animals like birds and insects excel in perceiving objects in rapid motion\cite{b32}. Taking inspiration from this phenomenon, event cameras are designed to detect changes in light intensity. Among them, the Dynamic Vision Sensor (DVS) is a representative event-based sensor that naturally detects moving objects and filters out redundant information automatically\cite{b1}. Unlike conventional RGB cameras that capture image frames consisting of the light intensity of every pixel, DVS cameras measure per-pixel brightness changes and output events asynchronously.  In the remainder of this paper, we refer to these events as visual events. The DVS cameras outperform their traditional counterparts in several aspects\cite{b1}. For instance, the microsecond temporal resolution of DVS cameras allows them to avoid motion blur, and their high dynamic range (140dB compared to the 60dB of conventional cameras) enables effective functioning in low illumination conditions.  They are particularly appealing for their low energy dissipation and are commonly used in energy-constrained systems where situation awareness is required. The visual events generated by the DVS camera are typically sparse. When processed by event-driven neuromorphic computing models, this sparsity allows for further reduction in computation energy.

Existing neuromorphic computing utilizes Spiking Neural Network (SNN) models, which are known as the third generation of neural networks. Spiking neurons communicate and process information through discrete electrical impulses referred to as spikes. Information is encoded using the rate or interval of these spikes. Similarly to biological neurons, SNNs are inherently asynchronous, and process information in an event-driven manner \cite{b32}.  SNNs have been successfully applied to  various tasks \cite{b14}\cite{b15}\cite{b16}\cite{b28}, demonstrating a potential in achieving high accuracy with low power consumption.  They have been utilized for classification {\cite{b45}\cite{b14}} and object detection {\cite{b46}\cite{b47}} of DVS camera output, enabling real-time visual analysis.

A DVS camera is connected to the back-end processor via a USB port. State of the art neuromorphic processors such as IBM TrueNorth \cite{b40} and Intel Loihi {\cite{b41}\cite{b42}} exhibit power consumption well below 1 Watt. The power consumption of a DVS camera can also be as low as 30mW{\cite{b43}}. The energy efficiency of data communication over USB channel ranges from approximately \(3.1\times10^{-8}\)  to \(1.4\times10^{-7}\) Joules per bit {\cite{b44}}. With a 128x128 image resolution, a conventional camera capturing 30 frames per second generates 107,520 pixels per second. Assuming that only 10\% of the pixels experience brightness change and it requires 3 bytes to describe an event, the DVS camera will generate approximately 258kb data per second. Consequently the transmission power required to move the data amounts to approximately 8 mW to 36 mW. Compared to the power consumption of the DVS camera and back-end processor, the power consumption of their data interface is not negligible.


 It is widely accepted that human brain functions as a predictive machine\cite{b29}\cite{b30}. The predictive coding model {\cite{b38}} proposes that the brain consistently generates and updates a "mental model" of the environment to anticipate input signals from the senses. The predictive behavior contributes to the exceptional energy efficiency of the human brain. Neuroscience experiments have demonstrated that neural activity decreases when a stimulus is expected to occur {\cite{b39}}. In other words, when the environment is predictable, the brain suppresses sensory inputs and relies on prior knowledge to infer information about the surrounding world. This phenomenon is known as expectation suppression.

 Inspired by the predictive coding and expectation suppression, we propose a predictive attention model to control the output of the DVS camera. The model consists of two main components: a predictor that anticipates future visual events and an attention generator that produces a control signal to regulate the output of the DVS sensor. The attention should be guided by the high-level goal of the application. In this study, our objective is to maintain a situation awareness, so attention is generated to ensure the accuracy of predicted visual events.

 To implement the visual event predictor, we employ an SNN-ANN hybrid auto-encoder. The SNN-based encoder models the feedforward neural pathway, while the artificial neural network (ANN) based decoder represents the feedback neural pathway. Depending on the attention level, the predictor receives inputs from either the DVS camera or its own predictions. When attention is high, the sensed events are utilized, whereas when attention is low, the predicted events are used. The predicted visual events can be seen as an internal belief about the world, therefore, we consider improving the prediction accuracy as equivalent to enhancing situation awareness.

The conventional approach to assess the quality of video prediction is by calculating the mean-square error (MSE) between the predicted frame and the ground truth frame \cite{b12} \cite{b13}. However, this metric is insufficient for measuring the similarity between two sets of visual events. In this study, we introduce a novel metric called event similarity (Esim) and  demonstrate that the Esim score demonstrates a stronger correlation with human visual perception compared to MSE.  

Our contributions can be summarized as follows:
\begin{itemize}
    \item We present an attention-assisted hybrid architecture that combines SNN and ANN for continuous visual event prediction. Compared to an ANN-based prediction model, our hybrid model on average improves the quality of prediction by $14.5\%$.
    \item We have developed a novel evaluation metric to assess the similarity between two sets of visual events. Unlike metrics such as MSE, our Esim measurement provides enhanced robustness against random noises and focuses more on the differences in salient features.
    \item The proposed attention model effectively gates $46.7\%$ of the communication between the DVS camera and the back-end processor, while maintaining an acceptable level of situation awareness. Compared to random gating at the same rate, a system utilizing attention-based gating achieves a $81.1\%$ higher level of situation awareness. 
    \item Moreover, by incorporating the predicted visual event, which contains significantly less noise and emphasizes salient features, we can reduce computation in the SNN-based encoder network by $43.8\%$, leading to more energy savings.  \end{itemize}

\section{BACKGROUND AND RELATED WORKS}

\subsection{DVS data representation}

A DVS camera generates an event when the logarithmically-scaled change in light intensity exceeds a certain threshold:
\begin{equation}
  \begin{cases}
    pos\ event\ \ log(I_{t+1})-log(I_{t})>th.\\
    neg\ event\ \ log(I_{t})-log(I_{t+1})>th.
  \end{cases}
\end{equation}
The event recording process is pixel-specific, where each pixel is evaluated independently of others. Each event consists of a 2D location of the event $(x, y)$, a timestamp \(t\), and a polarity \(p\) indicating whether it is a positive (light intensity increase) or negative event (light intensity decrease). Therefore, an event is recorded as a four-element tuple, $e=\left \{ x,y,t,p \right \}$, which is commonly known as Address Event Representation (AER).

Typically, visual events are grouped into event frames. Two grouping strategies are often used, i.e., dividing frames based on equally time interval or equal number of  events. In this study, we employ the first strategy. Each event frame is represented as a 2-dimensional ternary array, $ F\in B^{H\times W}$, where $B\in \left \{  0,1, -1 \right \}$ corresponds to no event, positive or negative events respectively. The dimension \(H \times W\) represents the camera resolution.  All events within the same frame are generated at approximately the same time, so their individual timestamps are disregarded during processing.

\subsection{Prediction Models}

Video prediction has been extensively studied in the field of deep learning. The objective is to generate a sequence of future frames, denoted as $\mathbf{Y}=(\hat{Y}_{t+1},\hat{Y}_{t+2},...,\hat{Y}_{t+m})$ given a sequence of $n$ preceding frames $\mathbf{X}=(X_{t-n},...,X_{t-1},X_{t})$ \cite{b10}. Training the video sequence model is self-supervised, as it does not require additional label information. 

To predict future frames, a model needs to possess a strong understanding of the spatiotemporal correlations within a video. Models such as recurrent neural networks (RNN) and Long Short-Term Memory (LSTM) \cite{b8}\cite{b2}\cite{b4}\cite{b3} are commonly used. While the outcome of a prediction model could be stochastic , the majority of works provide deterministic predictions \cite{b10}. The MSE metric is often used to evaluate the prediction quality. However, when dealing with event-based data, this method exhibits limitations due to the extreme sparsity of hot pixels (i.e., pixels with positive or negative events). Under the MSE metric, most frames appear "similar" to each other since large regions are often devoid of events. To the best of our knowledge, no prior works have been designed specifically to predict event frames from a DVS camera.

\subsection{Attention mechanism}

The attention mechanism has demonstrated significant success in machine learning, leading to improved model performance \cite{b19}\cite{b21}\cite{b22} and reduced computation resources \cite{b20}\cite{b22}. Generally, there are three types of attention mechanisms. Spatial-wise attention coupled with channel-wise attention is commonly used in convolutional neural network models\cite{b22},  while recurrent neural networks often employ temporal attention\cite{b23}. Although relatively new,  the attention mechanism has also been successfully applied in SNNs. For example, in \cite{b24}, SNN spatial attention is used for object recognition, while \cite{b27} focuses on discovering more informative period of events in a redundant DVS event stream using a squeeze-and-excitation\cite{b19} method. However, the soft attention employed in these works does not help reduce computation complexity. In contrast, our model generates hard attention, ensuring that the input is queried only when necessary, thus reducing communication and computation overhead.

\section{METHOD}


The overall architecture of the proposed visual event attention system comprises two main components: an autoencoder-based visual event predictor and an attention generator. The back-end processor buffers incoming visual events and combines them into event frames. The predictor takes an input event frame and predicts the next event frame. The attention generator evaluates the quality of the prediction and generates a binary gating signal to either block or unblock the camera output. The autoencoder takes a combination of the camera-sensed events and the predicted events as its input.

\begin{figure}[htb]
    \centering
    \includegraphics[scale=0.1]{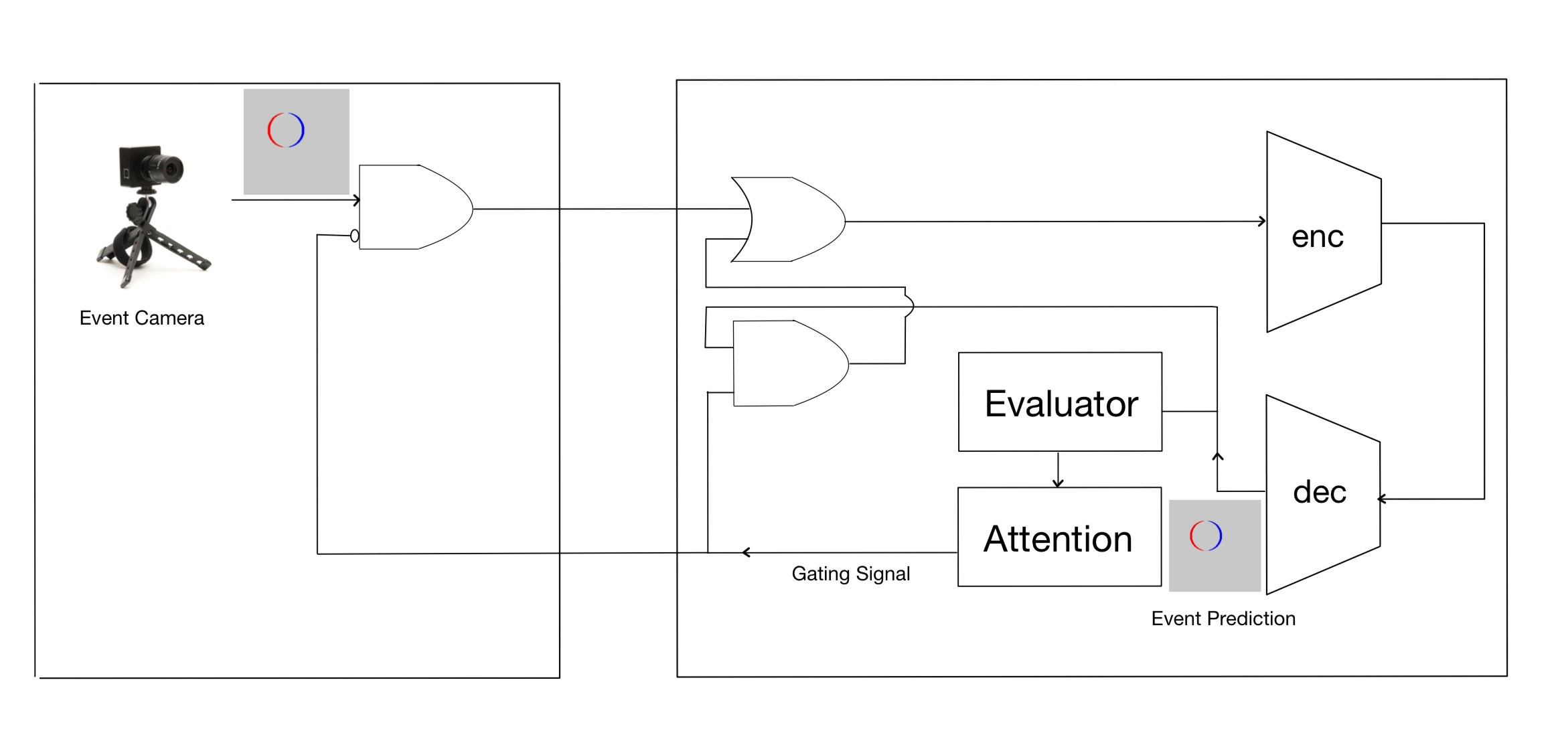}
    \caption{Overall visual predictive attention system architecture. 
    }
    \label{fig:attention}
\end{figure}

\subsection{Visual Event Predictor}
\begin{figure*}[htb]
    \centering
    \includegraphics[scale=0.23]{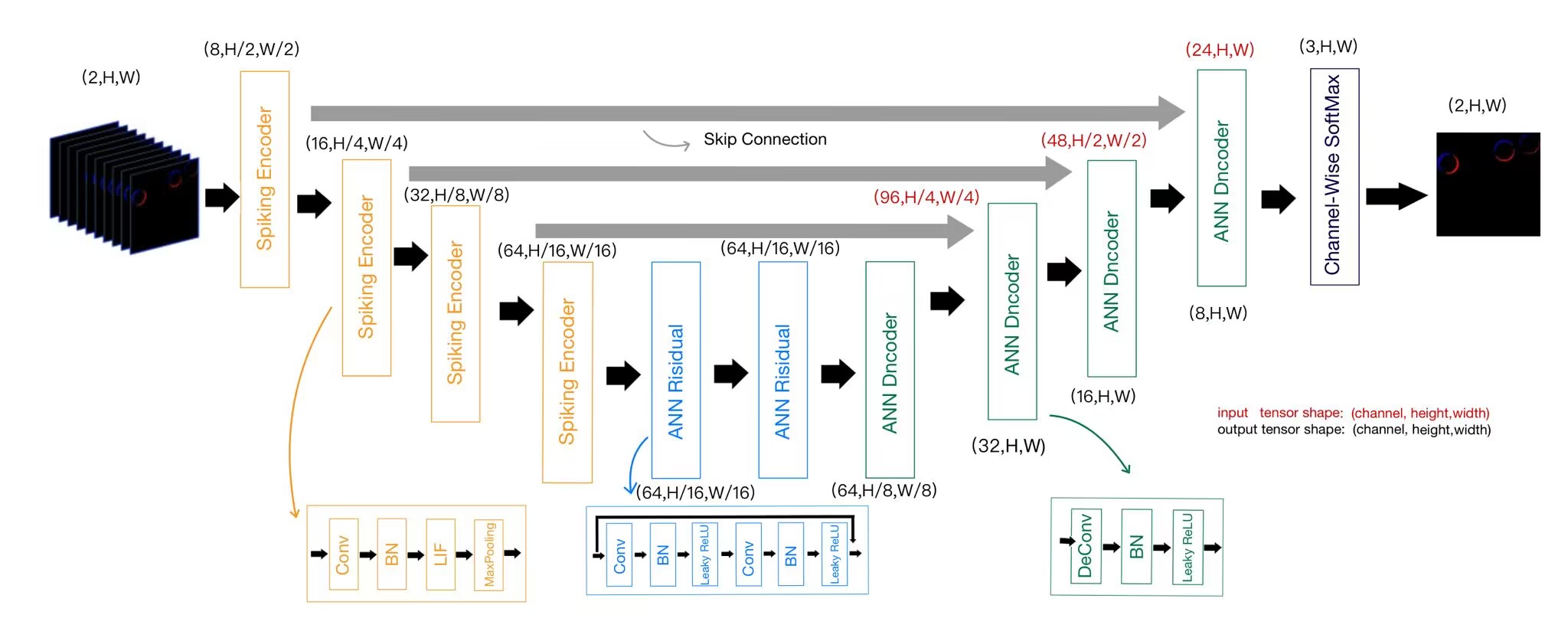}
    \caption{Visual event predictor. 
    }
    \label{fig:predictor}
\end{figure*}
The architecture of the visual event predictor is illustrated in Fig ~\ref{fig:predictor}. The predictor consists of three components: an SNN-based encoder, a sequence of residual blocks, and a decoder for next frame generation. Each layer of the encoder is a convolutional network of LIF neurons. The encoder extracts temporal and spatial features from the incoming sequence of event frames. The residual blocks are ANN-based convolutional layer with skip structure.  The decoder consists of multiple layers of ANN-based deconvolutional networks for generating the next frame. Feedforward skip connections are inserted between corresponding layers in the encoder and decoder. The output dimensions of each layer are labeled beside the blocks. While the SNN encoder leverages the sparsity in the DVS camera output, the ANN-based forecasting and frame generation provide more flexibility and higher trainability as the neurons output continuous values. Through experimentation, we discovered that the skip connection is crucial for achieving accurate predictions. Without the feedforward skip information, the sparse activities within the network often result in an all-zero output. Each pixel in the predicated frame may be in 3 possible states, i.e. positive, negative or no-event, a softmax layer is used at the output to predict the probability of the three possible states. Finally, events are sampled based on the probability.

We employ leaky-integrate-and-fire (LIF) neuron in the SNN encoder. The LIF neuron maintains a membrane potential \(u\) through leaky integration of input spikes. When the membrane potential exceeds the threshold, the neuron generates an output spike and resets the membrane potential. The dynamics of the LIF neuron are described by Equations  ~\ref{eq:2} and ~\ref{eq:3}. Here $u^{t,n} $ represents the membrane potential of a neuron in layer \(n\) at time $t$, $y^{t,n}$ denotes the neuron's output, and $V_{th}$ denotes the threshold of the neuron's membrane potential. In a convolutional-based SNN, each neuron is solely connected to and receives signals from other neurons within the receptive fields $F$ from the previous layer. 

\begin{equation} \label{eq:2}
    u^{t,n}=\displaystyle\sum\limits_{j\in F} w_{j}^{n}y_{j}^{t,n-1}+\tau u^{t-1,n} (1-y^{t-1,n})
\end{equation}

\begin{equation} \label{eq:3}
    y^{t,n}= \begin{cases} 
    1\text{ if } u^{t,n}>V_{th} \\
    0\text{ otherwise}
    \end{cases}
\end{equation}

The predictor is trained using Cross Entropy(CE) loss: 

\begin{equation}
    \mathcal{L}=-\frac{1}{H\cdot W} \displaystyle\sum\limits_{i=0}^{H\cdot W} \sum\limits_{j=0}^2 {y}_{i,j}\log_{}\widehat{y}_{i,j},
\end{equation}

where $\widehat{y}_{i,j}$ and ${y}_{i,j}$ represent the estimated and ground truth probability of pixel $i$ having event $j$. 

The activation function of the spiking neuron (Equation \ref{eq:3}) is non-differentiable, making it unsuitable for direct application of conventional backpropagation for training. A commonly used approach is to employ a surrogate function that approximates the gradient of the spikes' activation function. In this study, we utilize the arctangent function and its derivative as surrogates for calculating the derivative of the SNN activation function. The expressions for these surrogates are provided below, where $x$ and $y$ epresent the input and output of the activation function, respectively:
\begin{equation} 
    y =\frac{1}{\pi} \arctan(\frac{\pi}{2}\alpha x) + \frac{1}{2}
\end{equation}
\begin{equation} 
    \frac{\partial y}{\partial x} = \frac{\alpha}{2(1 + (\frac{\pi}{2}\alpha x)^2)}
\end{equation}

when $\alpha$ becomes larger, the function resembles a activation function for spiking neuron, for our experiment, we set $\alpha = 2 $.

\begin{figure}[htb]
\centering  
\subfigure[sensor input]{
\label{Fig.sub.1}
\includegraphics[width=3.2cm,height = 2.8cm]{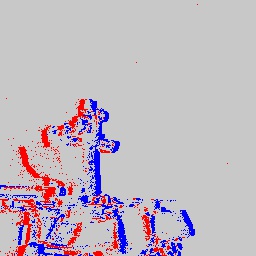}} 
\subfigure[model prediction]{
\label{Fig.sub.2}
\includegraphics[width=3.2cm,height = 2.8cm]{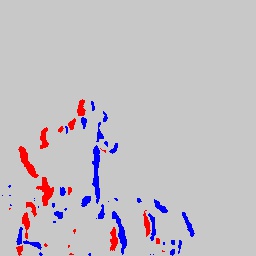}}
\caption{The prediction can reduce the noise and reflect a region activity}
\label{prediction-result}
\end{figure}

Fig ~\ref{prediction-result} presents an example of the predicted frame and the ground truth frame. In the image, the red and blue dots represent positive and negative events, respectively. Interestingly, we observed that the predicted frame exhibits significantly less noise compared to the original frame. This can be attributed to the fact that noise events are random and lack any discernible pattern, making them inherently unpredictable.

\subsection{Measuring Event Similarity}

 A reliable measure of similarity between two event frames should accurately capture the level of overlap in their visual events. Taking inspiration from the Intersection over Union (IOU) metric commonly used for evaluating object bounding boxes, we propose a novel metric called Event Similarity (Esim). Given two event frames, denoted as $F_{1}$ and $F_{2}$ each comprising positive and negative visual events, we begin by defining the intersection of $F_{1}$ and $F_{2}$ as $F_{1}\bigcap F_{2} =\left \{  e=\left \{ x,y,p \right \} \mid e\in F_{1}, e\in F_{2}  \right \}$. Furthermore, we define the union of $F_{1}$ and $F_{2}$ as $F_{1}\bigcup F_{2} =\left \{  e=\left \{ x,y,p \right \} \mid e\in F_{1} \parallel   \in F_{2} \right \} $. The intersection of two event frames only includes events that are present in both frames, while the union contains events from either of the frames. Esim is defined by the following equation:
 

\begin{equation}
    Esim(F_{1},F_{2})=\frac{\mid F_{1} \cap F_{2}\mid}{\mid F_{1} \cup F_{2}\mid}
\end{equation}
where $\mid \cdot\mid$ is $\ell{1}$ norm. Formula denotes the ratio of common visual events over the number of total visual events in these two frames.

Using Esim to assess the quality of predictions by comparing them to the ground truth frame has certain limitations.  Firstly, if the predicted event shifts from its ground truth location, regardless of the extent of the shift, there is no intersection between the two and the Esim score will be 0. Furthermore, due to its high sensitivity, the output of a DVS camera often contains a considerable number of noise events, sometimes even surpassing the number of informative events. Consequently, discrepancies with noise events can significantly diminish the Esim score of the predicted frame. 

The aforementioned limitations arise from the fact that Esim relies on exact event matching at a per-pixel level, which is unnecessary.  To introduce some tolerance towards random noise and small shifts, we compare the events within a small region surrounding the pixel.  Similar to using polarized filter for image noise reduction, we define the polarity intensity($PI$) of a ${h}\times{w}$ region centered at locate $(x,y)$ as
\begin{equation}
    PI(x,y)=\frac{|E_{pos}| - |E_{neg}| }{h\cdot w}
\end{equation}
where $|E_{pos}|$ and $|E_{neg}|$ represent the number positive and negative events within the specified region. Each pixel is polarized by checking the polarity intensity of its surrounding area against a threshold using the following equation:

\begin{equation}
P'(x, y)=\begin{cases}1 & PI(x, y) > th\\
-1 & PI(x, y) < -th \\0 & PI(x, y) < \lvert th\rvert \end{cases}
\end{equation}

The Esim calculated based on the array with polarized value of the pixel is refer to as Region Esim. Region Esim with a $n\times{n}$ polarization size is denoted as Esim$n$. The threshold can be tuned under different scenario(we use 0.2501, which proves to be fine value for our experiment for Esim2 and Esim4). Because one frame consists of random noise while the other does not. Despite featuring similar events related to ball movement, when using Esim, their similarity is nearly zero. However, when employing Region Esim, their similarity score exceeds to a reasonable level.

\begin{figure}[htb]
    \centering
    \includegraphics[scale=0.14]{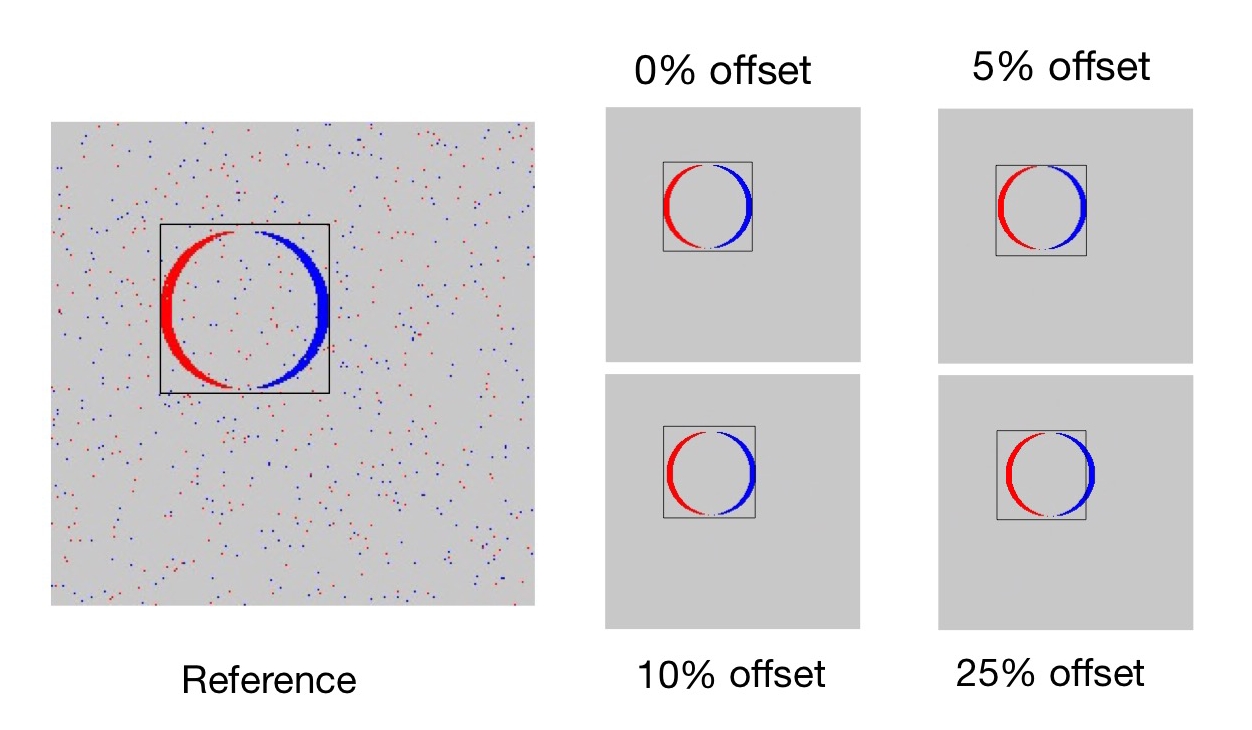}
    \caption{The prediction can reduce the noise and reflect a region activity. The left most figure is the reference, others are figures where the ball was moved to the right by a percentage of radius(the black box denotes the reference position).}
    \label{offset}
\end{figure}

We compared Esim with Mean Square Similarity (MSS), which is mathematically defined as $1-MSE$. The reference frame and the comparison frames are provided in Fig. ~\ref{offset}, with the white box indicating the location of the ball in the reference frame. The metric scores are reported in Table \ref{metric_compare}. It is evident that MSS fails to accurately depict the magnitude of the displacement of the ball's location. In contrast, Esim-based metrics is more suitable in measuring the similarity between the two event frames.

\begin{table}[ht]
\caption{Metric Score on Noised Shifted Ball}
\centering

\begin{tabular}{ |p{1.3cm}||p{1cm}|p{1cm}|p{1cm}|p{1cm}|}

 \hline offset & $0\%$  & $5\%$  & $10\%$ & $25\%$\\
 \hline
1-MSE  & 0.9949  & 0.9898 & 0.9848 & 0.9798\\
Esim     & 0.599  &0.3321 & 0.1438 & 0.0022\\
Esim2  & 0.9719  &0.5230 & 0.2433 & 0.0000\\
Esim4  & 0.9882  &0.5980 & 0.3412 & 0.0000\\
 \hline

\end{tabular}
  \label{metric_compare}
\end{table}

\subsection{Prediction Evaluator and attention generator}

Video prediction exhibits robust spatial and temporal consistency.  By feeding the predicted frame back into the autoencoder's input, we can continuously generate a sequence of future frames, extending beyond just the immediate next frame.  However, as this iterative process continues, errors accumulate, leading to a degradation in the prediction quality.   At this time, it becomes crucial to "take a look" (or pay attention) to the sensor stream once again in order to perceive the actual information and rectify the prediction errors. The challenge lies in determining when to look. 

As discussed in previous section, the region Esim can be used to measure the similarity between the predicted and the ground truth frame. However, since the system lacks access to the ground truth without "looking" at the camera output, we must estimate the Region Esim score based on the predicted frame.  Fortunately, this is feasible because a poorly generated event frame usually is noisy and the events do not form a clear contour of the object. By examining the event distribution within the predicted frame, we can make a rough estimation of how closely the prediction resembles the ground truth. Moreover, the quality of predictions often deteriorates significantly when the object changes its current trajectory. By observing the sequence of event frames, we can make a general prediction of when such a change is likely to occur.

\begin{figure}[htb]
    \centering
    \includegraphics[scale=0.12]{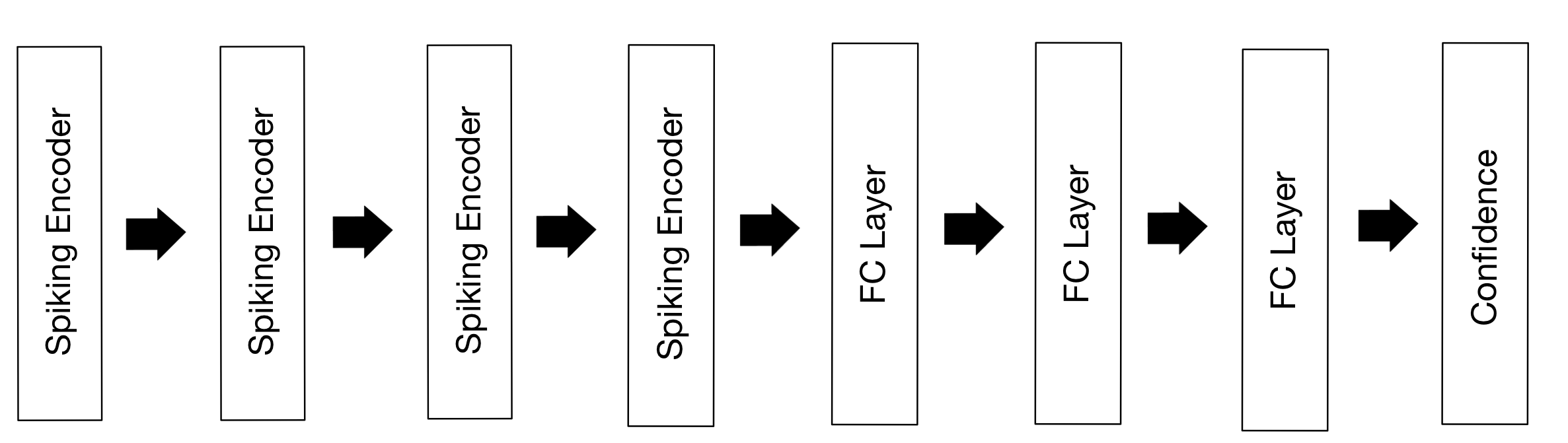}
    \caption{Esim estimation model}
    \label{Esim_estimation_model}
\end{figure}

In this work, we developed a neural network based evaluator to predict the Region Esim score of the predicted frame. The architecture of the evaluator is illustrated in figure ~\ref{Esim_estimation_model}. The evaluator architecture is a typical SNN feature extractor. It takes channel-wise concatenation of the first frame and the current prediction as input and the concatenation was continuously fed into the evaluator for 10 time steps and then outputs a confidence of current prediction. The training of the evaluator is same as and independent from the predictor.

\section{Experiment Result}

\subsection{Experimental Setup}
Our experiments were conducted on four datasets: BouncingBall, MovingMnist, Visevent\cite{b9}, and DVS128\cite{amir2017low}. All of them contain sequences of event frames from a DVS camera. Both BouncingBall and MovingMnist are synthetic datasets. BouncingBall offers resolutions of 64x64 or 256x256 and features three balls moving in straight path and bouncing off boundaries or each other upon collision.  MovingMnist, on the other hand, has a resolution of 64x64 and showcases two MNIST digits bouncing within the scene. Visevent is specifically designed for object tracking of pedestrians and vehicles in street scenarios. DVS128 captures the upper body's moving gestures of a man, with a resolution of $128\times128 $. For each dataset, we use 50\% of the sequences as training set and the remaining as the testing set. During the training, the learning rate is set to $1e-3$, the batch size to $8$, and a total of $ 2000 $ epochs were trained .

\subsection{Predictor Performance}

\begin{table}[ht]
\caption{ Predictor Performance}
\centering

\begin{tabular}{ |p{2.0cm}||p{1.3cm}|p{0.8cm}|p{1.6cm}|}
 \hline
Dataset & frame size & SNN &  ANN \\
 \hline
 BouncingBall64  & (64,64)   & 0.801 & 0.7289\\
 BouncingBall256 & (256,256) & 0.589 & 0.5124\\
 MovingMnist     & (64,64)   & 0.691 & 0.6702\\
 DVSGesture      &(128,128)  & 0.502 & 0.3918\\
 Visevent        &(256,256)  & 0.317 & 0.2726\\
 \hline

\end{tabular}
  \label{predictor_performance}
\end{table}

We tested our predictor on aforementioned datasets and reported the average Esim4 score in Table ~\ref{predictor_performance}. 
We also trained an ANN predictor as a baseline. It has a similar architecture except that an LSTM-based encoder is used to replace the SNN encoder. However, due to the very sparse input, the accuracy of ANN prediction is much lower than SNN. In addition, the ANN in general does not support event-driven operation, hence will have more computation workload.

\begin{figure}[htb]
\centering
    \includegraphics[scale=0.1]{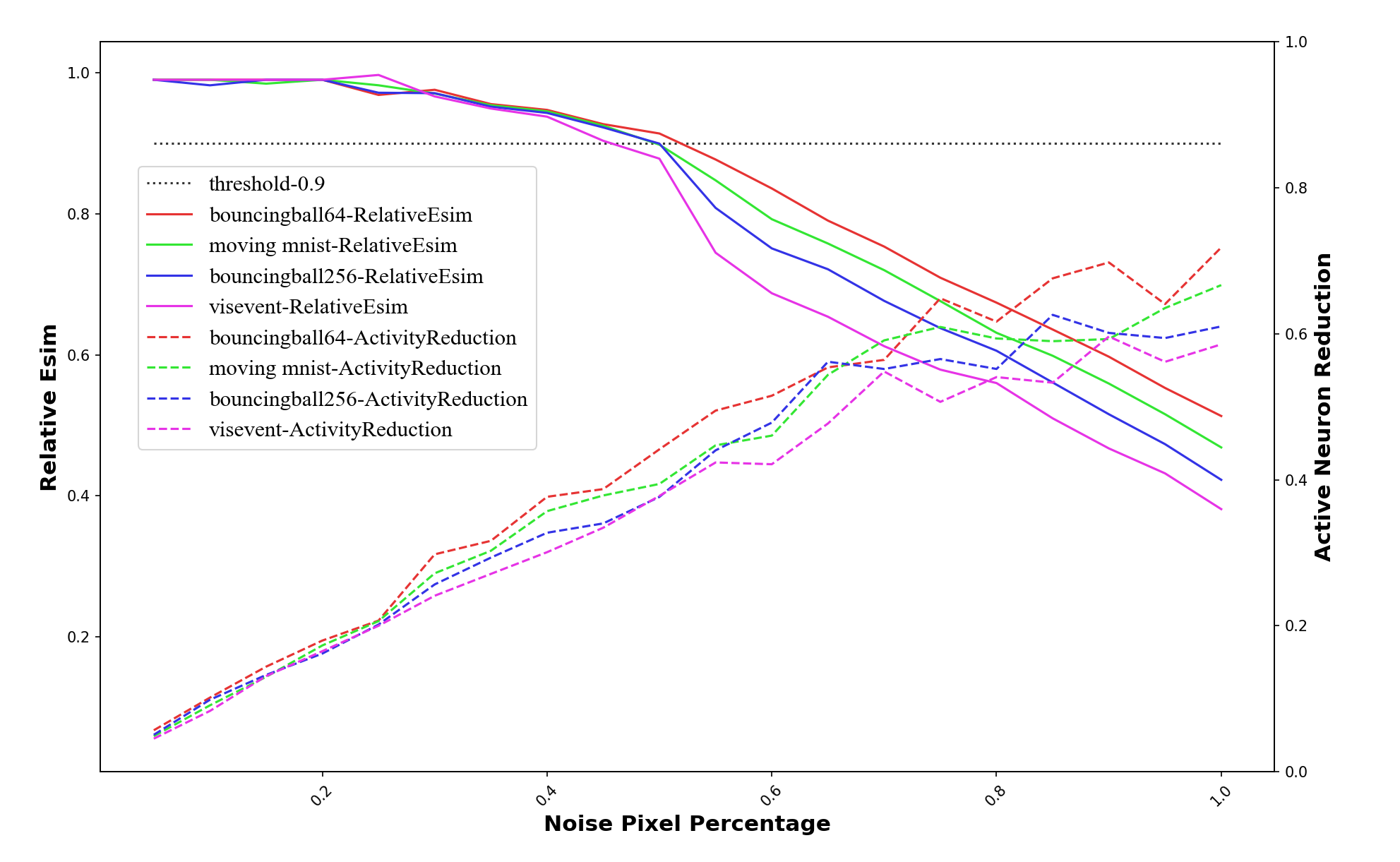}
    \caption{Relative performance reduction as noise increases.}
    \label{relative_performance}
\end{figure}

The predictor also demonstrates its ability to significantly reduce noise in the input. To evaluate its noise tolerance, we manually added Gaussian noise to the test set and passed the noisy frames through the predictor. We introduced a term called "relative Esim" to quantify the prediction's robustness. The relative Esim is calculated as the ratio of the Esim scores between predictions with noisy input and predictions with clean input. By gradually increasing the noise level, we observed the corresponding changes in the relative Esim. The trends for the four datasets are depicted in Fig \ref{relative_performance}. As we can see, the predictor can tolerate about from $47.3\%$ (bouncingball64) to $52.1\%$ (visevent) noise, while maintain a prediction that is $90\%$ similar to the prediction with clean input. In other words, the predictor is capable of filtering out from $32.0\%$(bouncingball64) to $43.8\%$ (visevent)noisy events. This can be turned into reduced computation in the event-driven SNN encoder.

\subsection{Attention Directed Situation Awareness}

\begin{figure}[htb]
    \centering
    \includegraphics[scale=0.09]{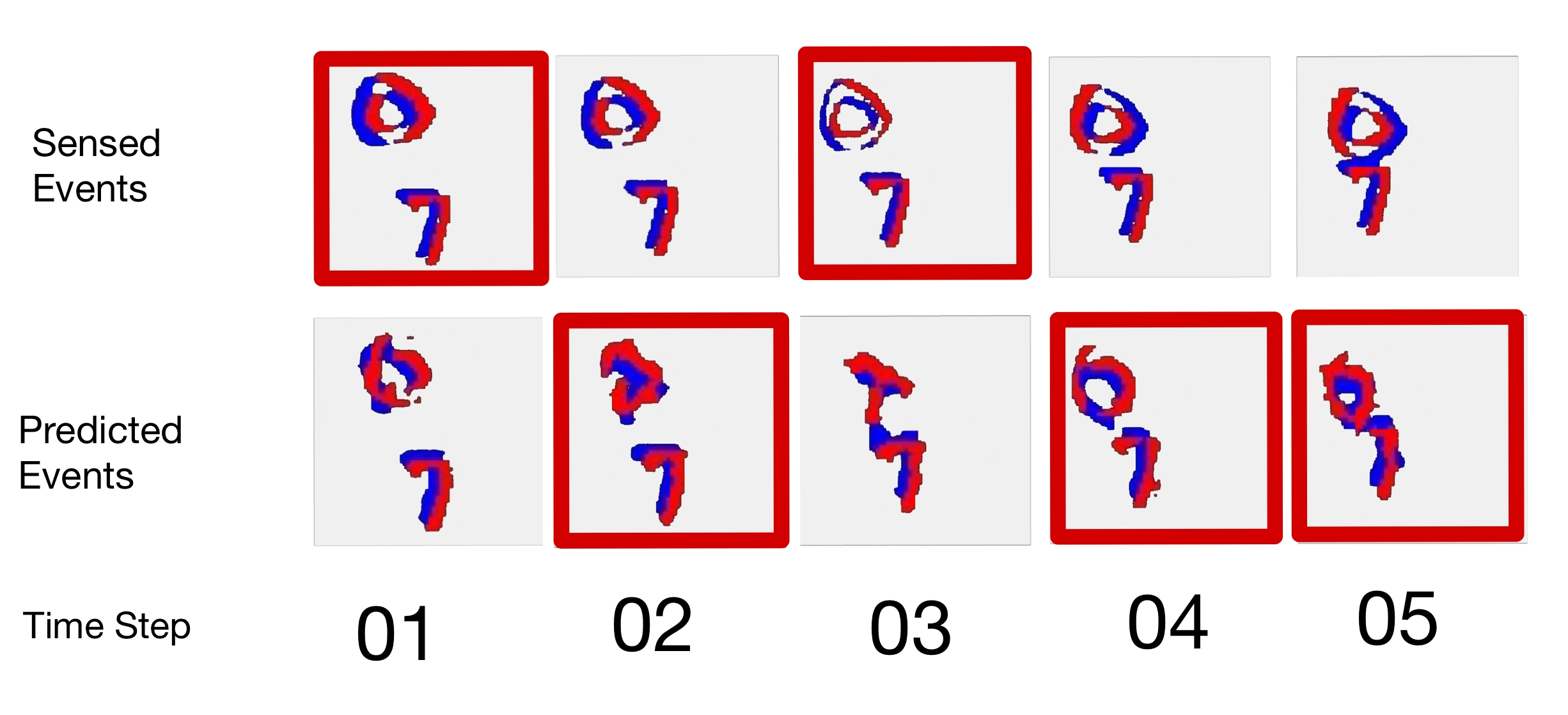}
    \caption{Attention assisted sequence prediction of MovingMnist (The red box indicates the input of next time step.) }
    \label{mm07}
\end{figure}

 We further apply the evaluator to continuously estimate the Esim score of the prediction. If the score falls below a threshold, the camera output is attended, and a sensed frame is used to replace the poorly predicted frame while the prediction continues. Fig \ref{mm07} shows an example of predicted frames. The 3th frame is replaced by the sensed information.

\begin{figure}[htb]
    \centering
    \includegraphics[scale=0.17]{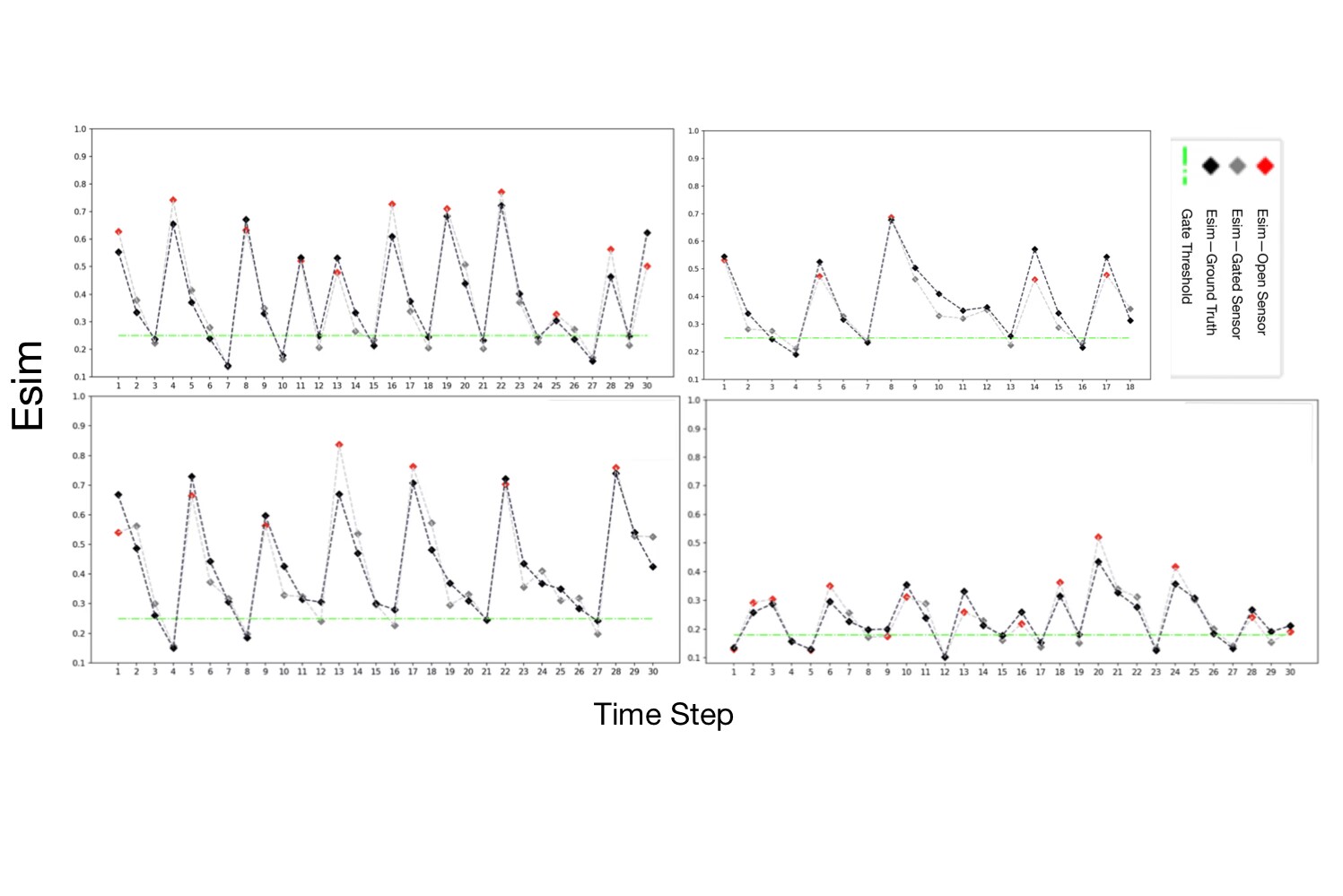}
    \caption{Evaluator assisted Situation Awareness.}
    \label{assissted_awareness}
\end{figure}

Fig ~\ref{assissted_awareness} illustrates how the prediction quality varies throughout the attention process. The black lines represent the actual Esim score of the prediction after comparing it with the ground truth, while the grey lines represent the Esim score estimated by the evaluator.  The red dot indicates the predicted Esim right after the sensor is attended, and the green line represents the threshold. As we can see that the evaluator gives quite good estimation of the Esim score of the prediction. Each time the sensor input is received, the prediction quality is reinforced. We can also see that, when the predictor maintaing an acceptable relative performance, with the proposed attention mechanism, the sensor processor interface can be gated for about {76.7\%}(bouncingball64) of time on average, leading to energy reduction in data communication. Even in the most complex scenario(visevent), the gating rate can still be up to {46.7\%} . Due to the fact that the prediction contains significantly fewer noise events compared to the sensor input, when the sensor is gated, the encoder experiences lower computational activity. This observation aligns with the neurological observation that neurons exhibit reduced activity when attention levels are low.

\begin{figure}[htb]
    \centering
    \includegraphics[scale=0.12]{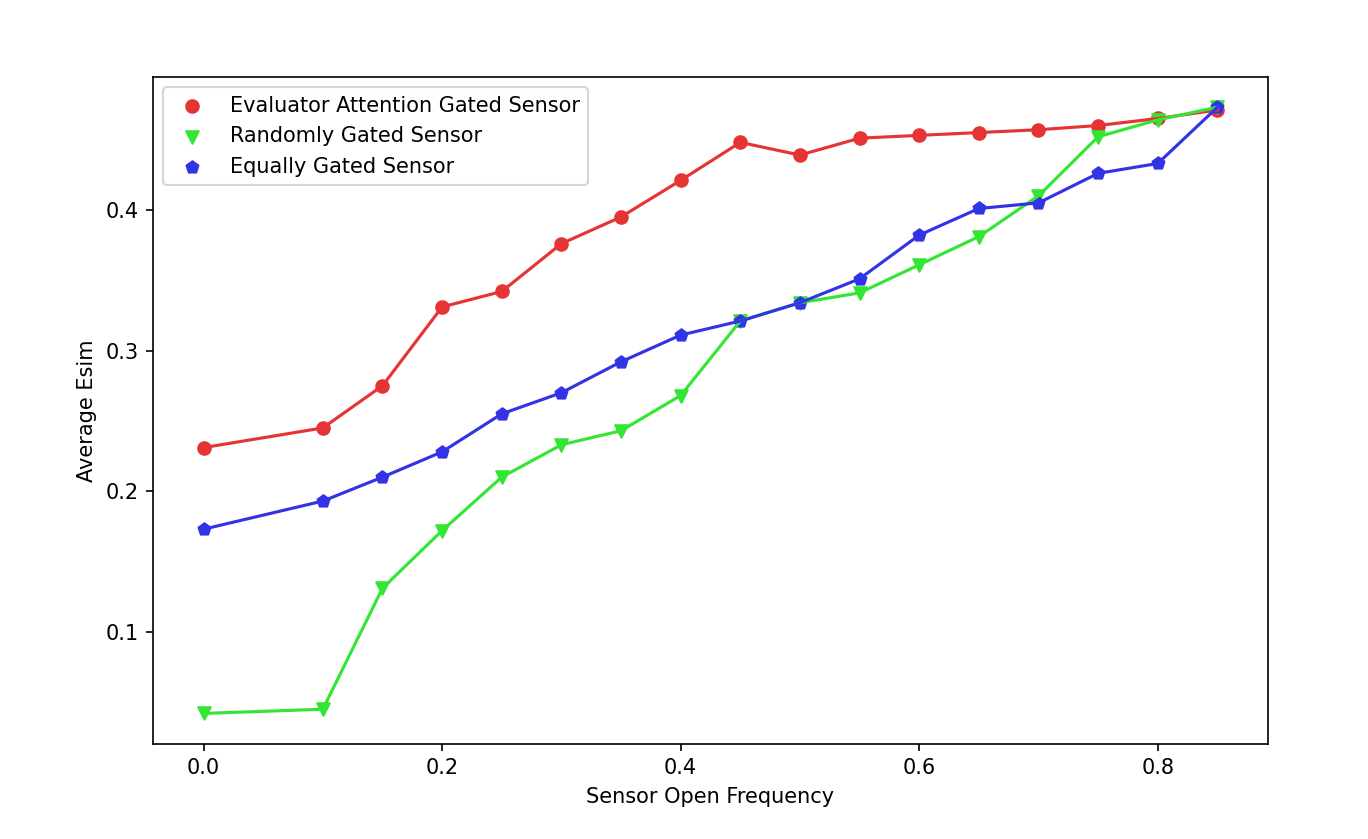}
    \caption{Average score comparison to random gated sensor and periodically gated sensor.}
    \label{Compared_to_random_gating}
\end{figure}

We further show that the attention is carefully timed to ensure situation awareness. To evaluate this, we compared our attention mechanism with a randomly and a periodicallytimed attention mechanism  using the same attention rate. By adjusting the evaluation threshold, we varied the rough attention rate, which represents the percentage of time that the sensor is open, ranging from 90\% to 10\%. We used the quality of the prediction as a metric to measure situation awareness. The results are presented in  Fig. ~\ref{Compared_to_random_gating}. It is evident that, on average, our attention model provides 81\% better awareness compared to the random attention mechanism.

\subsection{Conclusion}

In this work, we have designed a generative model for predicting the next frame using inputs from a DVS camera. This model combines the strengths of a deep SNN encoder and an ANN decoder. We have introduced Esim as an evaluation metric to assess the quality of the predictions. An evaluator is developed to estimate the Esim score from predicted event frames and controls the output of the DVS camera based on the attention level. We have compared the performance of the predictive model with both the proposed attention mechanism and a random attention mechanism. The results demonstrate that the proposed attention mechanism achieves superior situation awareness. Additionally, the predicted frames contain significantly less noise, resulting in reduced computational requirements. Overall, the proposed predictive attention mechanism offers a promising approach to decrease both communication and computational energy consumption.

\section*{Acknowledgement}
This work is partially supported by the National Science
Foundation I/UCRC ASIC (Alternative Sustainable and Intelligent Computing) Center (CNS-1822165).
\bibliographystyle{IEEEtran}
\bibliography{root}

\begin{thebibliography}{10}
\providecommand{\url}[1]{#1}
\csname url@samestyle\endcsname
\providecommand{\newblock}{\relax}
\providecommand{\bibinfo}[2]{#2}
\providecommand{\BIBentrySTDinterwordspacing}{\spaceskip=0pt\relax}
\providecommand{\BIBentryALTinterwordstretchfactor}{4}
\providecommand{\BIBentryALTinterwordspacing}{\spaceskip=\fontdimen2\font plus
\BIBentryALTinterwordstretchfactor\fontdimen3\font minus \fontdimen4\font\relax}
\providecommand{\BIBforeignlanguage}[2]{{%
\expandafter\ifx\csname l@#1\endcsname\relax
\typeout{** WARNING: IEEEtran.bst: No hyphenation pattern has been}%
\typeout{** loaded for the language `#1'. Using the pattern for}%
\typeout{** the default language instead.}%
\else
\language=\csname l@#1\endcsname
\fi
#2}}
\providecommand{\BIBdecl}{\relax}
\BIBdecl

\bibitem{b32}
M.~F. Land, ``Motion and vision: why animals move their eyes,'' \emph{Journal of Comparative Physiology A}, vol. 185, pp. 341--352, 1999.

\bibitem{b1}
G.~Gallego, T.~Delbr{\"u}ck, G.~Orchard, C.~Bartolozzi, B.~Taba, A.~Censi, S.~Leutenegger, A.~J. Davison, J.~Conradt, K.~Daniilidis \emph{et~al.}, ``Event-based vision: A survey,'' \emph{IEEE transactions on pattern analysis and machine intelligence}, vol.~44, no.~1, pp. 154--180, 2020.

\bibitem{b14}
\BIBentryALTinterwordspacing
M.~Yao, H.~Gao, G.~Zhao, D.~Wang, Y.~Lin, Z.~Yang, and G.~Li, ``Temporal-wise attention spiking neural networks for event streams classification,'' \emph{CoRR}, vol. abs/2107.11711, 2021. [Online]. Available: \url{https://arxiv.org/abs/2107.11711}
\BIBentrySTDinterwordspacing

\bibitem{b15}
\BIBentryALTinterwordspacing
D.~Roy, P.~Panda, and K.~Roy, ``Synthesizing images from spatio-temporal representations using spike-based backpropagation,'' \emph{CoRR}, vol. abs/1906.08861, 2019. [Online]. Available: \url{http://arxiv.org/abs/1906.08861}
\BIBentrySTDinterwordspacing

\bibitem{b16}
\BIBentryALTinterwordspacing
H.~Kamata, Y.~Mukuta, and T.~Harada, ``Fully spiking variational autoencoder,'' \emph{CoRR}, vol. abs/2110.00375, 2021. [Online]. Available: \url{https://arxiv.org/abs/2110.00375}
\BIBentrySTDinterwordspacing

\bibitem{b28}
\BIBentryALTinterwordspacing
Y.~Hu, J.~Binas, D.~Neil, S.~Liu, and T.~Delbr{\"{u}}ck, ``{DDD20} end-to-end event camera driving dataset: Fusing frames and events with deep learning for improved steering prediction,'' \emph{CoRR}, vol. abs/2005.08605, 2020. [Online]. Available: \url{https://arxiv.org/abs/2005.08605}
\BIBentrySTDinterwordspacing

\bibitem{b45}
\BIBentryALTinterwordspacing
X.~She, S.~Dash, and S.~Mukhopadhyay, ``Sequence approximation using feedforward spiking neural network for spatiotemporal learning: Theory and optimization methods,'' in \emph{The Tenth International Conference on Learning Representations, {ICLR} 2022, Virtual Event, April 25-29, 2022}.\hskip 1em plus 0.5em minus 0.4em\relax OpenReview.net, 2022. [Online]. Available: \url{https://openreview.net/forum?id=bp-LJ4y\_XC}
\BIBentrySTDinterwordspacing

\bibitem{b46}
H.~Liu, D.~P. Moeys, G.~Das, D.~Neil, S.-C. Liu, and T.~Delbrück, ``Combined frame- and event-based detection and tracking,'' in \emph{2016 IEEE International Symposium on Circuits and Systems (ISCAS)}, 2016, pp. 2511--2514.

\bibitem{b47}
D.~Tedaldi, G.~Gallego, E.~Mueggler, and D.~Scaramuzza, ``Feature detection and tracking with the dynamic and active-pixel vision sensor (davis),'' in \emph{2016 Second International Conference on Event-based Control, Communication, and Signal Processing (EBCCSP)}, 2016, pp. 1--7.

\bibitem{b40}
F.~Akopyan, J.~Sawada, A.~Cassidy, R.~Alvarez-Icaza, J.~Arthur, P.~Merolla, N.~Imam, Y.~Nakamura, P.~Datta, G.-J. Nam \emph{et~al.}, ``Truenorth: Design and tool flow of a 65 mw 1 million neuron programmable neurosynaptic chip,'' \emph{IEEE transactions on computer-aided design of integrated circuits and systems}, vol.~34, no.~10, pp. 1537--1557, 2015.

\bibitem{b41}
M.~Davies, N.~Srinivasa, T.-H. Lin, G.~Chinya, Y.~Cao, S.~H. Choday, G.~Dimou, P.~Joshi, N.~Imam, S.~Jain, Y.~Liao, C.-K. Lin, A.~Lines, R.~Liu, D.~Mathaikutty, S.~McCoy, A.~Paul, J.~Tse, G.~Venkataramanan, Y.-H. Weng, A.~Wild, Y.~Yang, and H.~Wang, ``Loihi: A neuromorphic manycore processor with on-chip learning,'' \emph{IEEE Micro}, vol.~38, no.~1, pp. 82--99, 2018.

\bibitem{b42}
\BIBentryALTinterwordspacing
G.~Orchard, E.~P. Frady, D.~B.~D. Rubin, S.~Sanborn, S.~B. Shrestha, F.~T. Sommer, and M.~Davies, ``Efficient neuromorphic signal processing with loihi 2,'' \emph{CoRR}, vol. abs/2111.03746, 2021. [Online]. Available: \url{https://arxiv.org/abs/2111.03746}
\BIBentrySTDinterwordspacing

\bibitem{b43}
\BIBentryALTinterwordspacing
M.~Zaffar, S.~Ehsan, R.~Stolkin, and K.~D. McDonald{-}Maier, ``Sensors, {SLAM} and long-term autonomy: {A} review,'' \emph{CoRR}, vol. abs/1807.01605, 2018. [Online]. Available: \url{http://arxiv.org/abs/1807.01605}
\BIBentrySTDinterwordspacing

\bibitem{b44}
\BIBentryALTinterwordspacing
S.~Lurye, ``{Surges in mobile energy consumption during USB charging and data exchange – Securelist},'' 7 2016. [Online]. Available: \url{https://securelist.com/surges-in-mobile-energy-consumption-during-usb-charging-and-data-exchange/75297/}
\BIBentrySTDinterwordspacing

\bibitem{b29}
A.~Clark, ``Whatever next? predictive brains, situated agents, and the future of cognitive science,'' \emph{Behavioral and brain sciences}, vol.~36, no.~3, pp. 181--204, 2013.

\bibitem{b30}
T.~P. Lillicrap, A.~Santoro, L.~Marris, C.~J. Akerman, and G.~Hinton, ``Backpropagation and the brain,'' \emph{Nature Reviews Neuroscience}, vol.~21, no.~6, pp. 335--346, 2020.

\bibitem{b38}
Y.~Huang and R.~P. Rao, ``Predictive coding,'' \emph{Wiley Interdisciplinary Reviews: Cognitive Science}, vol.~2, no.~5, pp. 580--593, 2011.

\bibitem{b39}
K.~S. Walsh and D.~P. McGovern, ``Expectation suppression dampens sensory representations of predicted stimuli,'' \emph{Journal of Neuroscience}, vol.~38, no.~50, pp. 10\,592--10\,594, 2018.

\bibitem{b12}
Z.~Wang, A.~Bovik, H.~Sheikh, and E.~Simoncelli, ``Image quality assessment: from error visibility to structural similarity,'' \emph{IEEE Transactions on Image Processing}, vol.~13, no.~4, pp. 600--612, 2004.

\bibitem{b13}
Y.~Zhou, H.~Dong, and A.~El~Saddik, ``Deep learning in next-frame prediction: A benchmark review,'' \emph{IEEE Access}, vol.~8, pp. 69\,273--69\,283, 2020.

\bibitem{b10}
S.~Oprea, P.~Martinez-Gonzalez, A.~Garcia-Garcia, J.~A. Castro-Vargas, S.~Orts-Escolano, J.~Garcia-Rodriguez, and A.~Argyros, ``A review on deep learning techniques for video prediction,'' \emph{IEEE Transactions on Pattern Analysis and Machine Intelligence}, vol.~44, no.~6, pp. 2806--2826, 2022.

\bibitem{b8}
C.~Lee, A.~K. Kosta, A.~Z. Zhu, K.~Chaney, K.~Daniilidis, and K.~Roy, ``Spike-flownet: event-based optical flow estimation with energy-efficient hybrid neural networks,'' in \emph{Computer Vision--ECCV 2020: 16th European Conference, Glasgow, UK, August 23--28, 2020, Proceedings, Part XXIX 16}.\hskip 1em plus 0.5em minus 0.4em\relax Springer, 2020, pp. 366--382.

\bibitem{b2}
\BIBentryALTinterwordspacing
W.~Lotter, G.~Kreiman, and D.~D. Cox, ``Deep predictive coding networks for video prediction and unsupervised learning,'' \emph{CoRR}, vol. abs/1605.08104, 2016. [Online]. Available: \url{http://arxiv.org/abs/1605.08104}
\BIBentrySTDinterwordspacing

\bibitem{b4}
\BIBentryALTinterwordspacing
Y.~Wang, Z.~Gao, M.~Long, J.~Wang, and P.~S. Yu, ``Predrnn++: Towards {A} resolution of the deep-in-time dilemma in spatiotemporal predictive learning,'' \emph{CoRR}, vol. abs/1804.06300, 2018. [Online]. Available: \url{http://arxiv.org/abs/1804.06300}
\BIBentrySTDinterwordspacing

\bibitem{b3}
\BIBentryALTinterwordspacing
R.~Villegas, J.~Yang, Y.~Zou, S.~Sohn, X.~Lin, and H.~Lee, ``Learning to generate long-term future via hierarchical prediction,'' \emph{CoRR}, vol. abs/1704.05831, 2017. [Online]. Available: \url{http://arxiv.org/abs/1704.05831}
\BIBentrySTDinterwordspacing

\bibitem{b19}
\BIBentryALTinterwordspacing
J.~Hu, L.~Shen, and G.~Sun, ``Squeeze-and-excitation networks,'' \emph{CoRR}, vol. abs/1709.01507, 2017. [Online]. Available: \url{http://arxiv.org/abs/1709.01507}
\BIBentrySTDinterwordspacing

\bibitem{b21}
\BIBentryALTinterwordspacing
A.~Vaswani, N.~Shazeer, N.~Parmar, J.~Uszkoreit, L.~Jones, A.~N. Gomez, L.~Kaiser, and I.~Polosukhin, ``Attention is all you need,'' \emph{CoRR}, vol. abs/1706.03762, 2017. [Online]. Available: \url{http://arxiv.org/abs/1706.03762}
\BIBentrySTDinterwordspacing

\bibitem{b22}
\BIBentryALTinterwordspacing
S.~Woo, J.~Park, J.~Lee, and I.~S. Kweon, ``{CBAM:} convolutional block attention module,'' \emph{CoRR}, vol. abs/1807.06521, 2018. [Online]. Available: \url{http://arxiv.org/abs/1807.06521}
\BIBentrySTDinterwordspacing

\bibitem{b20}
\BIBentryALTinterwordspacing
A.~Papadopoulos, P.~Korus, and N.~D. Memon, ``Hard-attention for scalable image classification,'' \emph{CoRR}, vol. abs/2102.10212, 2021. [Online]. Available: \url{https://arxiv.org/abs/2102.10212}
\BIBentrySTDinterwordspacing

\bibitem{b23}
\BIBentryALTinterwordspacing
I.~Sutskever, O.~Vinyals, and Q.~V. Le, ``Sequence to sequence learning with neural networks,'' \emph{CoRR}, vol. abs/1409.3215, 2014. [Online]. Available: \url{http://arxiv.org/abs/1409.3215}
\BIBentrySTDinterwordspacing

\bibitem{b24}
\BIBentryALTinterwordspacing
M.~Cannici, M.~Ciccone, A.~Romanoni, and M.~Matteucci, ``Attention mechanisms for object recognition with event-based cameras,'' \emph{CoRR}, vol. abs/1807.09480, 2018. [Online]. Available: \url{http://arxiv.org/abs/1807.09480}
\BIBentrySTDinterwordspacing

\bibitem{b27}
\BIBentryALTinterwordspacing
M.~Yao, H.~Gao, G.~Zhao, D.~Wang, Y.~Lin, Z.~Yang, and G.~Li, ``Temporal-wise attention spiking neural networks for event streams classification,'' \emph{CoRR}, vol. abs/2107.11711, 2021. [Online]. Available: \url{https://arxiv.org/abs/2107.11711}
\BIBentrySTDinterwordspacing

\bibitem{b9}
X.~Wang, J.~Li, L.~Zhu, Z.~Zhang, Z.~Chen, X.~Li, Y.~Wang, Y.~Tian, and F.~Wu, ``Visevent: Reliable object tracking via collaboration of frame and event flows,'' \emph{arXiv preprint arXiv:2108.05015}, 2021.

\bibitem{amir2017low}
A.~Amir, B.~Taba, D.~Berg, T.~Melano, J.~McKinstry, C.~Di~Nolfo, T.~Nayak, A.~Andreopoulos, G.~Garreau, M.~Mendoza \emph{et~al.}, ``A low power, fully event-based gesture recognition system,'' in \emph{Proceedings of the IEEE conference on computer vision and pattern recognition}, 2017, pp. 7243--7252.

\end{thebibliography}

\end{document}